\documentclass[sigconf]{acmart}

\AtBeginDocument{%
  }

\copyrightyear{2022}
\acmYear{2022}
\setcopyright{rightsretained}
\acmConference[RecSys '22]{Sixteenth ACM Conference on Recommender Systems}{September 18--23, 2022}{Seattle, WA, USA}
\acmBooktitle{Sixteenth ACM Conference on Recommender Systems (RecSys '22), September 18--23, 2022, Seattle, WA, USA}
\acmDOI{10.1145/3523227.3547384}
\acmISBN{978-1-4503-9278-5/22/09}

\DeclareMathOperator*{\argmin}{arg\,min}
\DeclareMathOperator*{\argsort}{arg\,sort}
\usepackage{natbib}
\usepackage{subcaption}
\usepackage{wrapfig}
\usepackage{url}

\begin{document}

\title{Dynamic Surrogate Switching: Sample-Efficient Search \\ for Factorization Machine Configurations in Online Recommendations}

\author{Bla\v{z} \v{S}krlj}
\email{bskrlj@outbrain.com}
\orcid{1234-5678-9012}
\author{Adi Schwartz}
\email{aschwartz@outbrain.com}
\author{Jure Ferle\v{z}}
\email{jferlez@outbrain.com}
\author{Davorin Kopi\v{c}}
\email{dkopic@outbrain.com}
\author{Naama Ziporin}
\email{nziporin@outbrain.com}
\affiliation{%
  \institution{Outbrain Inc.}
  \country{USA}
}

\renewcommand{\shortauthors}{\v{S}krlj et al.}

\begin{abstract}
Hyperparameter optimization is the process of identifying the appropriate hyperparameter configuration of a given machine learning model with regard to a given learning task. For smaller data sets, an exhaustive search is possible; However, when the data size and model complexity increase, the number of configuration evaluations becomes the main computational bottleneck. A promising paradigm for tackling this type of problem is surrogate-based optimization. The main idea underlying this paradigm considers an incrementally updated model of the relation between the hyperparameter space and the output (target) space; the data for this model are obtained by evaluating the main learning engine, which is, for example, a factorization machine-based model. By learning to approximate the hyperparameter-target relation, the surrogate (machine learning) model can be used to score large amounts of hyperparameter configurations, exploring parts of the configuration space beyond the reach of direct machine learning engine evaluation. Commonly, a surrogate is selected prior to optimization initialization and remains the same during the search. We investigated whether dynamic switching of surrogates during the optimization itself is a sensible idea of practical relevance for selecting the most appropriate factorization machine-based models for large-scale online recommendation. We conducted benchmarks on data sets containing hundreds of millions of instances against established baselines such as Random Forest- and Gaussian process-based surrogates. The results indicate that surrogate switching can offer good performance while considering fewer learning engine evaluations. 
\end{abstract}



\maketitle
\section{Introduction}
\label{sec:intro}
Online recommendation is fundamental to functioning of the modern web. From online stores to streaming service engines and content discovery platforms, it enables users to discover relevant content faster, and product owners to make their products visible to a broader public. Recommender systems can be understood as a feedback loop between the recommendation engine and the target audience – based on provided recommendations, the interactions with the recommendations are stored and considered to improve a given recommender further~\cite{batmaz2019review}.

Machine learning models underlie many online recommendation systems. The most common branch of algorithms considered, both due to their performance and lower complexity, are \textbf{factorization machines (FM)}. These algorithms require the modelers to provide both the space of features and their interactions (\emph{featurization}), as well as model hyperparameters~\cite{rendle2012factorization}. The latter is the main focus of this paper. Tuning hyperparameters can be considered as an optimization task, formulated as follows:
$$\textrm{Solution} \approx \argmin_{\substack{\Theta \in  \textrm{hyperParamSpace}}}  \mathbb{E} \big [\textsc{Loss} \big ( \textrm{\emph{LearningEngine}},\Theta,\textrm{data}\big ) \big ],$$
\noindent where \emph{data} corresponds to the training data of choice, $\Theta$ to a given hyperparameter configuration, and \emph{LearningEngine} to the machine learning model considered. The minimization aims to identify minimal expected \emph{Loss} of choice (lower is better in this formulation). This formulation considers one \emph{LearningEngine} evaluation for each configuration. As the optimization progresses, configuration-target pairs are obtained and stored. \emph{Surrogate models} exploit this data to estimate which configuration the \emph{LearningEngine} will consider next. The problem with larger data sets is that the amount of engine evaluations is \emph{limited}, hence having \textbf{data-efficient surrogates} is a desired property that can substantially shorten research cycles. The contributions presented are multi-fold, and are stated next.
First, we describe the process of integration of surrogate-based optimization into the \textbf{in-house AutoML framework} -- we considered both Bayesian and non-Bayesian (surrogate) models, reporting on their behaviour.
We next present the implemented strategy of \textbf{Dynamic Surrogate Switching} (DSS) that builds on the recent ideas of automated model switching during optimization.
The results of initial evaluations indicate this is a promising strategy to speed up hyperparameter optimization in real-life settings, where the number of learning engine evaluations is limited due to the scale of data considered.
\section{Selected Related work}
\label{sec:related}

\begin{figure}
 \centering
    \Description[Example 3D visualization of hyperparameter landscape.]{Hyperparameters visualized: two hyperparameters with corresponding RIG scores.}
    \includegraphics[width=1\linewidth]{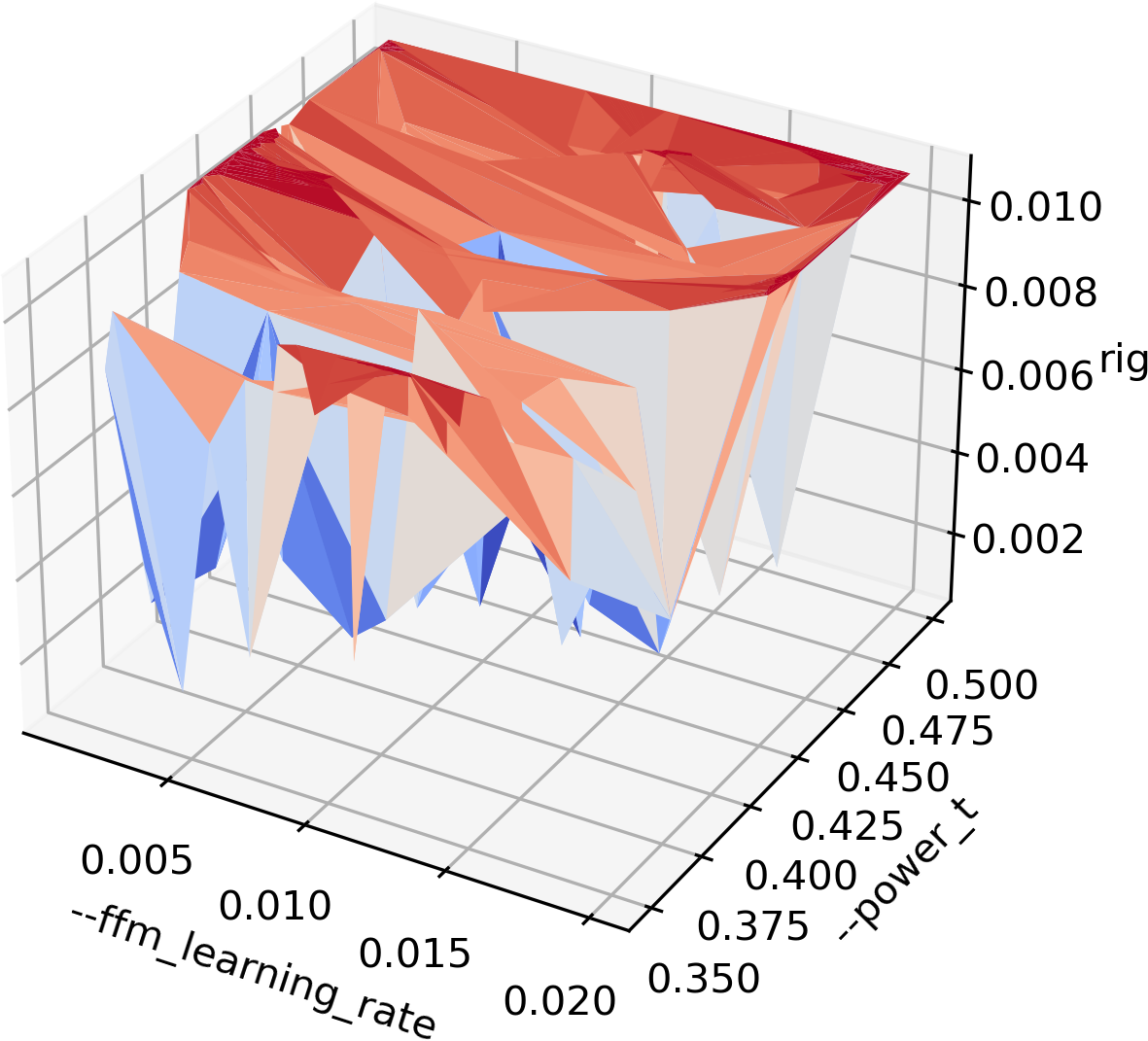}
    \caption{Example two-hyperparameter non-convex (interpolated) landscape.}
    \label{fig:nonconvex}
\end{figure}

The vast majority of machine learning models require the specification of \textbf{hyperparameters} for their normal mode of operation. Early approaches mostly considered grid-based, random, or evolution-based search through the space of possible configurations~\cite{friedrichs2005evolutionary}. However, recent trends indicate that, albeit random search often represents a viable baseline, exploiting the information obtained during the optimization can be a better strategy. There are multiple ways these hyperparameter-target score data points can be utilized. The simplest examples include incorporation of their statistical properties into the search strategy, however, considering them as input data samples for the \emph{surrogate model} is also often considered~\cite{falkner2018bohb}. The ideas of \textbf{surrogate-based} optimization are tightly linked with the field of Bayesian optimization~\cite{malu2021bayesian}; effectively, if a surrogate is a probabilistic model capable of also outputting the (epistemic) uncertainty associated with a given prediction, both the prediction and the associated uncertainty can be used by the \emph{acquisition function} -- the procedure responsible for linking the surrogate's output(s) with parts of the search space that should be considered in the next round of learning engine evaluations. There exist a plethora of commonly considered acquisition functions -- examples include the probability of improvement, expected improvement and more~\cite{frazier2018tutorial}. If the surrogate is only capable of outputting the prediction, acquisition functions are simpler -- for example, viable configurations can be identified already by sorting the space of candidate configurations according to the surrogate's outputs. The \textbf{acquisition function} is thus responsible for using a trained surrogate to \emph{score} parts of the hyperparameter space -- these can be obtained via random sampling or more involved optimization schemes (such as for example, using L-BFGS~\cite{zhu1997algorithm} to minimize the acquisition function directly). Non-convex optimization landscapes can emerge already when considering as few as two hyperparameters in production environments (example shown in Figure~\ref{fig:nonconvex}).
The presented methodology also builds on the recent ideas of Online \textbf{AutoML} (Automated Machine Learning)~\cite{he2021automl}. The goal of AutoML systems is to automate various data-related processes commonly manually performed by human modelers. In particular, we built on some of the ideas presented recently in~\cite{celik2022online}, where dynamic ensembles were used to perform online learning. The main finding of the aforementioned study is that \emph{switching} the models can help mitigate problems related to data quantity (at the initial stages) and concept drift (at the latter stages). Albeit the considered setting is being tested offline, we can interpret the search itself as a dynamic process -- commonly, a single surrogate type is considered throughout the search, however, is this the preferred strategy? We explored whether dynamic re-configuration of surrogate models \emph{during the optimization} is a sensible, sample-efficient hyperparameter optimization strategy suitable for \textbf{large-scale} model configuration search.

\begin{figure}
 \centering
       \Description[Workflow depicting DSS.]{A cyclic workflow showing how the models are iteratively refined/switched.}
    \includegraphics[width=0.95\linewidth]{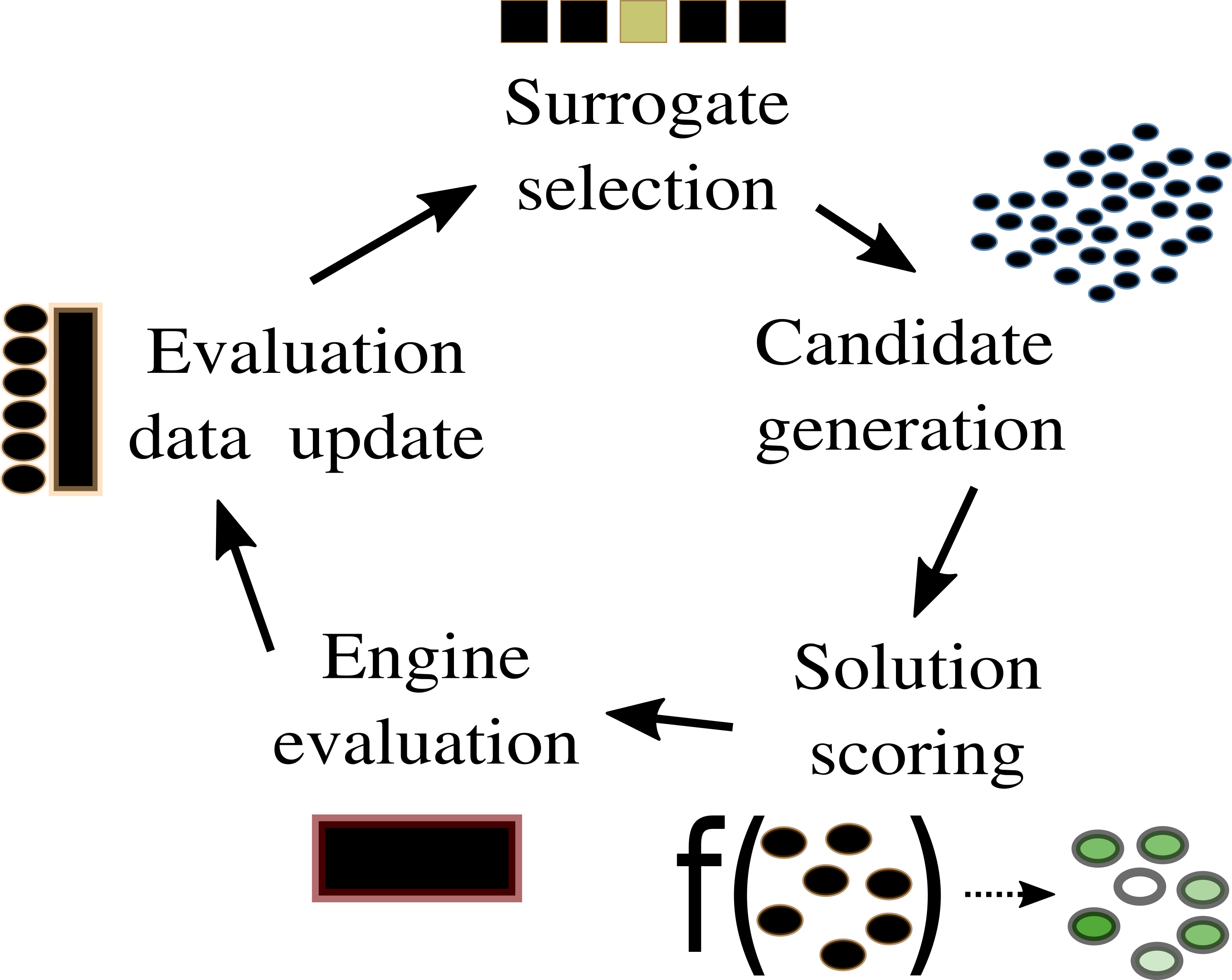}
    \caption{Conceptual overview of \textbf{DSS}, part of the in-house AutoML.}
    \label{fig:scheme}
\end{figure}

\begin{figure}[t!]
     \centering
      \Description[Two figures showing DSS's performance]{Left image shows iterative model switching in action, right one shows benchmark performance against strong baselines.}
     \begin{subfigure}[b]{0.45\textwidth}
         \centering
         \includegraphics[width=\textwidth, height=4.3cm]{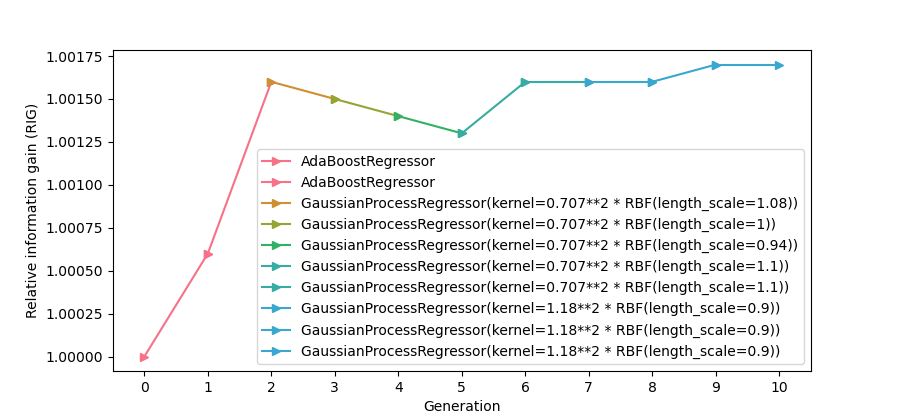}
         \caption{Visualizing Dynamic Surrogate Switching -- the surrogates change as the optimization progresses.}
     \end{subfigure}
     \hfill
     \begin{subfigure}[b]{0.45\textwidth}
         \centering
         \includegraphics[width=\textwidth, height=4cm]{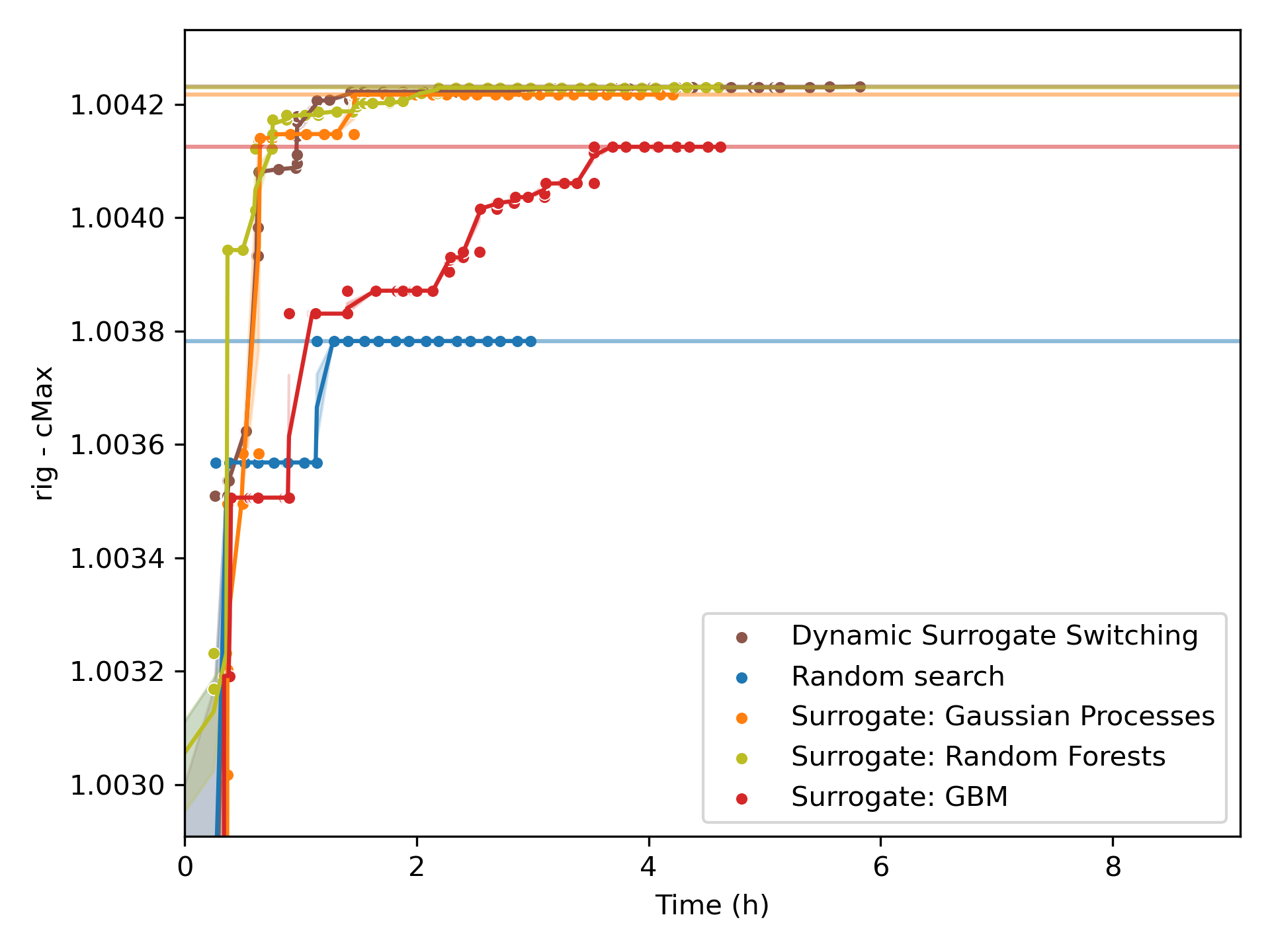}
         \caption{Benchmark results -- offline tests -- CTR task (Relative information gain~\cite{he2014practical}) with fixed stopping.}
     \end{subfigure}
        \caption{Visualization of DSS benchmark experiment and an ablation. The goal was to identify whether surrogate switching is a feasible strategy to obtain good model configurations. The benchmark results on a private data set consisting of hundreds of millions of instances (target was click-through rate -- CTR) are shown in Figure b). Results indicate DSS is a promising strategy candidate for sample-efficient hyperparameter optimization, requiring minimal human involvement during configuration.}
        \label{fig:three graphs}
\end{figure}

\section{Dynamic Surrogate Switching (DSS)}
\label{sec:method}
Surrogate models learn to estimate configuration quality during hyperparameter optimization. Changing surrogates during the search itself was previously shown to have a positive effect on the search's efficiency~\cite{Mehmani2018}. 
This work builds on similar ideas, extending them beyond common function optimization benchmark sets to operate as a part of the internal \emph{AutoML framework} that requires handling of \textbf{hundreds of millions} of instances during the search for suitable FM configurations.
This section describes the key components of the presented method and the initial study of their behaviour. We refer to the proposed method as Dynamic Surrogate Switching (DSS). A conceptual overview of the proposed method is shown in Figure~\ref{fig:scheme}. The approach follows the paradigm of surrogate-based modeling extended by the idea of surrogate switching. We continue with \textbf{an overview of DSS}.
As the first step, we will consider the evaluation of the learning engine (in our case, field-aware factorization machines (FFMs)~\cite{juan2016field}\footnote{Field aware factorization machines implementation available as \url{https://github.com/outbrain/fwumious_wabbit}.}. Initially, diverse parts of the search space are selected and considered by the learning engine to obtain the initial set of configuration-score tuples suitable for surrogate-based learning. The \textbf{evaluation data update} step is responsible for storing the configurations in a format suitable for learning. Further, this step also checks for anomalies in the evaluations (e.g., too many similar/same results). 
Note that the evaluation data set is constantly updated throughout the search, with all data considered each update (to maximize utilization of prior evaluations).
The subsequent step of \textbf{surrogate selection} is what differentiates DSS from conventional surrogate-based learning the most -- the DSS is based on the previous work~\cite{Mehmani2018} and the ideas of OAML, aimed to enable dynamic surrogate re-configuration. We implemented it by considering a collection of different model types and their initial configurations. The considered models that can serve as a surrogate are, for example, Random Forests~\cite{breiman2001random}, Gaussian Processes~\cite{rasmussen2003gaussian} and Gradient Boosting Machines~\cite{friedman2001greedy}~\footnote{Implemented with components from~\cite{varoquaux2015scikit}.}. As the evaluation of such surrogates is inexpensive (and thus not the bottleneck of the whole search), the surrogate selection phase considers multiple parametrizations of the mentioned model types -- overall, hundreds of surrogates are evaluated each iteration. The next issue we tackled was how to \emph{score} the surrogates. We perform the selection based on the \emph{proportion of explained variance}. Thus, the surrogate model scoring (including ranking) can be formulated as:
$\argsort_{s \in S}  \frac{\textrm{Var} (y - s(D))}{\textrm{Var} (y)},$
\noindent where $S$ is the set of possible surrogate models and $D$ the current set of configuration-score tuples obtained by the learning engine during search. The highest-ranked surrogate is selected and used during the acquisition step for a given search iteration.
An important component of surrogate-based optimization is \textbf{configuration generation}. 
Solution candidates are scored in mini-batches to tackle the memory overhead. We further augmented the random configuration search with a memory structure that discards parts of the space considered previously. Once the surrogate is selected, re-trained on the current data and the candidate space is selected, \textbf{solution scoring} takes place. Here, the surrogate is used to provide a score for each generated configuration. There are multiple possible ways of utilizing the obtained scores; e.g., in Bayesian search, probability-of-improvement or similar heuristics can be considered. As the considered DSS does not always offer probabilistic outputs, simpler acquisition functions are required. We exploit the fact that the AutoML system that implements DSS runs in \textbf{multi-threaded} mode, enabling us to devote some threads to top-ranked solutions obtained by the surrogate while leaving some threads to perform exploration based on more randomized configurations. Once selected, the learning engine considers the configurations, and the feedback loop continues. Example results are summarized in Figure~\ref{fig:three graphs}. Overall, the DSS was identified as a promising approach to tuning FFMs.

\section{Conclusions}
\label{sec:conclusions}
The paper presents Dynamic Surrogate Switching (DSS), an approach aimed to facilitate hyperparameter search -- one of the components of the in-house AutoML utilized daily by many data scientists. The results indicate that dynamic surrogates indeed offer a promising alternative to existing strong baselines (e.g., Random Forest-based surrogates), are potentially more adaptive to particular problems considered, and require less human intervention during configuration. Further work includes testing the idea on different data sets and studying its behaviour when considering different learning engines.

\section*{AUTHOR BIO}
\label{sec:bio}
Bla\v{z} \v{S}krlj is a machine learning researcher active in the areas of AutoML and representation learning. During his PhD at the Jo\v{z}ef Stefan International Postgraduate School he worked on low-resource AutoML and its applications to natural language processing, graph-based machine learning and compressibility of latent representations. Currently, Bla\v{z} is part of Outbrain’s AutoML team, where he is exploring the limits of AutoML for large-scale recommendation.

\bibliographystyle{ACM-Reference-Format}
\bibliography{sample-base}

\end{document}